%% file: main.tex
\documentclass[10pt,twocolumn,letterpaper]{article}

\usepackage[pagenumbers]{cvpr} % To force page numbers, e.g. for an arXiv version

\input{preamble}

\definecolor{cvprblue}{rgb}{0.21,0.49,0.74}
\usepackage[pagebackref,breaklinks,colorlinks,allcolors=cvprblue]{hyperref}

\usepackage[subtle]{savetrees} % uncomment to save space if necessary

\title{Sample- \textit{and} Parameter-Efficient Auto-Regressive Image Models}

\author{Elad Amrani\\
Apple, Technion\\
\and
Leonid Karlinsky\\
MIT-IBM Watson AI Lab\\
\and
Alex Bronstein\\
Technion
\and
{\small \url{https://github.com/elad-amrani/xtra}}
}
\begin{document}
\input{sec/promo_figure}
\input{sec/abstract}
\input{sec/intro}
\input{sec/method_figure}
\input{sec/related_work}
\input{sec/method}
\input{sec/implementation_details}
\input{sec/quantitative_results}
\input{sec/ablations}
\input{sec/discussion}

\input{sec/qualitative_results}

\input{sec/summary}

\clearpage
{
    \small
    \bibliographystyle{ieeenat_fullname}
    \bibliography{main}
}

\input{sec/suppl}

\end{document}

%% file: preamble.tex
\usepackage{graphicx}
\usepackage{amsmath}
\usepackage{amssymb}
\usepackage{booktabs}
\usepackage{colortbl}
\usepackage{xcolor}
\usepackage{arydshln}
\usepackage{float}    % For controlling float positions
\usepackage{afterpage}
\usepackage{caption}

\definecolor{lightgray}{rgb}{0.9, 0.9, 0.9}
\definecolor{lightgreen}{rgb}{0.88, 1, 0.88}

%% file: sec/promo_figure.tex
\twocolumn[{%
\renewcommand\twocolumn[1][]{#1}%
\maketitle
\begin{center}
    \centering
    \captionsetup{type=figure}
    \begin{subfigure}[t]{0.497\linewidth}
        \includegraphics[width=\linewidth]{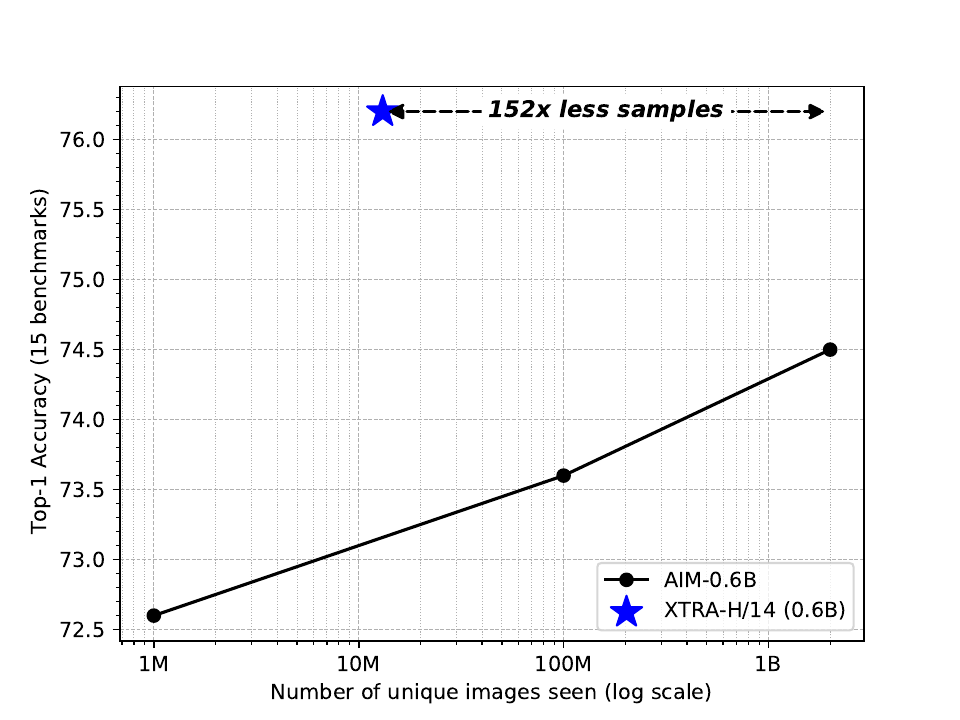}
    \end{subfigure}
    \hfill
    \begin{subfigure}[t]{0.497\linewidth}
        \includegraphics[width=\linewidth]{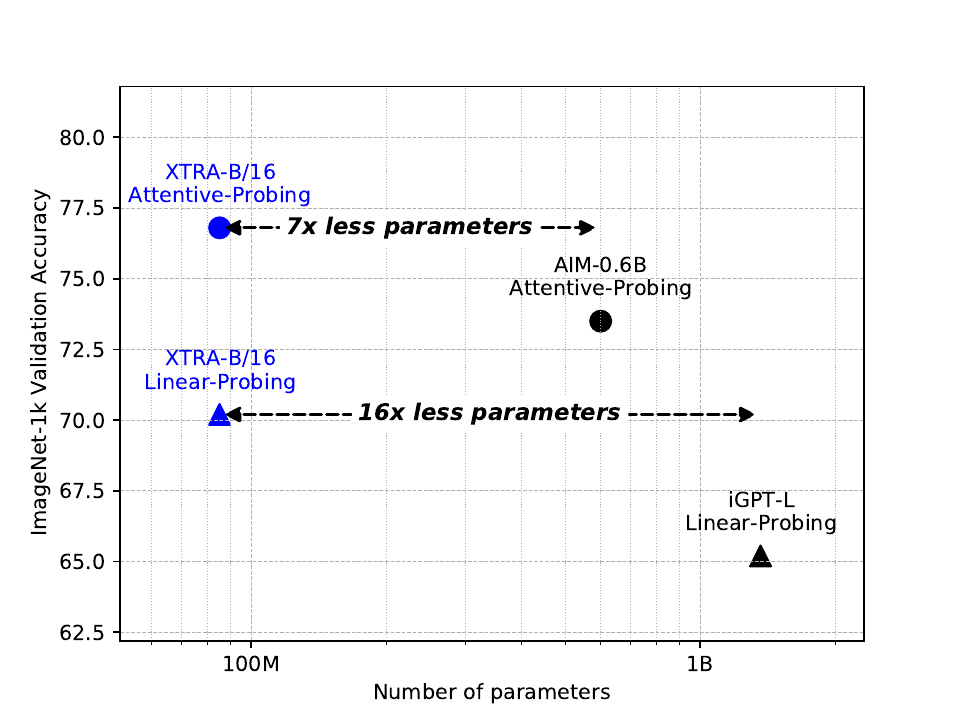}
    \end{subfigure}
    \caption{\textbf{Sample and Parameter Efficiency of XTRA.} (Left) XTRA-H/14 (0.6B parameters) outperforms prior state-of-the-art auto-regressive image model (AIM-0.6B \cite{el2024scalable}) in top-1 average accuracy across 15 diverse image recognition benchmarks, despite being trained on 152$\times$ fewer samples. (Right) XTRA-B/16 (85M parameters) outperforms prior auto-regressive image models trained on ImageNet-1k in linear and attentive probing tasks, while using 7–16$\times$ fewer parameters.}
    \label{fig:teaser_figure}
\end{center}
}]

%% file: sec/abstract.tex
\begin{abstract}
We introduce \textbf{XTRA}, a vision model pre-trained with a novel auto-regressive objective that significantly enhances both sample and parameter efficiency compared to previous auto-regressive image models. Unlike contrastive or masked image modeling methods, which have not been demonstrated as having consistent scaling behavior on unbalanced internet data,
auto-regressive vision models exhibit scalable and promising performance as model and dataset size increase. In contrast to standard auto-regressive models, XTRA employs a Block Causal Mask, where each Block represents $k \times k$ tokens rather than relying on a standard causal mask. By reconstructing pixel values block by block, XTRA captures higher-level structural patterns over larger image regions. Predicting on blocks allows the model to learn relationships across broader areas of pixels, enabling more abstract and semantically meaningful representations than traditional next-token prediction.
This simple modification yields two key results. First, \textbf{XTRA is sample-efficient}. Despite being trained on 152$\times$ fewer samples (13.1M vs. 2B), XTRA ViT-H/14 surpasses the top-1 average accuracy of the previous state-of-the-art auto-regressive model across 15 diverse image recognition benchmarks. Second, \textbf{XTRA is parameter-efficient}. Compared to auto-regressive models trained on ImageNet-1k, XTRA ViT-B/16 outperforms in linear and attentive probing tasks, using 7–16$\times$ fewer parameters (85M vs. 1.36B/0.63B).
\end{abstract}
\begingroup
\renewcommand\thefootnote{}\footnote{Elad Amrani conducted this work while at IBM Research-AI.}
\addtocounter{footnote}{0}
\endgroup

%% file: sec/intro.tex
\section{Introduction}
\label{sec:intro}
Auto-regressive models have played a foundational role in recent advancements in Natural Language Processing (NLP). The simplicity of their objective — predicting the next word in a sequence based on its preceding context — allows these models to capture intricate dependencies and complex patterns over long sequences. This next-token prediction mechanism has proven highly effective, leading to the development of models capable of sophisticated language understanding and generation. Most importantly, auto-regressive language models \cite{radford2019language,brown2020gpt3,touvron2023llama,touvron2023llama2,dubey2024llama3} demonstrate desirable scaling properties, where downstream performance improves consistently as both model capacity and data size grow \cite{kaplan2020scaling}.

In contrast, the progress in the field of Computer Vision (CV) has been driven mostly by Contrastive Learning (CL) methods \cite{dosovitskiy2014discriminative,chen2020simclr,wu2018unsupervised,he2020momentum,chen2020improved,chen2020big,caron2020unsupervised,chen2021exploring,caron2021emerging,amrani2022self}, which aim to learn visual representations by maximizing the similarity of two different augmentations of the same image while simultaneously minimizing the similarity between different images; Masked Image Modeling (MIM) methods \cite{bao2022beit,he2022masked,xie2022simmim,baevski2022data2vec,baevski2023efficient,assran2023self,chen2024context}, which focus on predicting masked parts of an image based on its visible regions; and various hybrid approaches that combine elements of both \cite{zhou2021ibot,oquab2023dinov2}. These methods, and specifically the hybrid approaches, set the current state-of-the-art performance for self-supervised visual representation learning. Despite their success, these methods often rely on intricate training recipes involving numerous tricks, such as multi-crop \textit{handcrafted} augmentations, momentum networks, schedules for teacher momentum and weight decay, and complex regularization techniques like KoLeo \cite{sablayrolles2018spreading} and LayerScale \cite{touvron2021going}. These modifications can introduce significant overhead, both in terms of computational resources and implementation complexity. While some methods, like DINOv2 \cite{oquab2023dinov2}, have shown promising scaling behavior, they lack the consistent scaling laws \cite{kaplan2020scaling} that are a hallmark of auto-regressive models. For example, \cite{singh2023effectiveness} demonstrated that, even when scaling the pre-training dataset from 1M to 3B samples using MAE ViT-H \cite{he2022masked}, the resulting improvement on downstream tasks is modest, with only a 0.5\% gain on ImageNet1k \cite{deng2009imagenet} and a 1.2\% gain on iNAT-18 \cite{inaturalist18}. This limited scaling effectiveness can impede these models' capacity to sustain performance gains as they increase in size or are trained on larger, uncurated internet datasets, in contrast to the more predictable scaling benefits observed in auto-regressive language models.

Auto-regressive image models \cite{chen2020generative,el2024scalable}, like their language counterparts, predict image pixels (or patches) sequentially based on the preceding context. These models aim to learn visual representations by capturing dependencies across pixel or patch sequences. First, iGPT \cite{chen2020generative} demonstrated the feasibility of self-supervised visual representation learning using an auto-regressive model that predicts the next pixel. Later, AIM \cite{el2024scalable}, a model based on Vision Transformers (ViT), demonstrated that auto-regressive models for images can scale similarly to their NLP counterparts, offering a consistent relationship between the model’s objective function and its downstream task performance. This characteristic is crucial for scalable visual representation learning, as it enables predictable improvements as model size or dataset size increases. However, despite their promising scaling properties, both iGPT and AIM suffer from significant sample- and parameter-efficiency drawbacks. iGPT, for example, required 7 billion parameters to achieve results on par with contrastive models that operate with 20 times fewer parameters. Similarly, AIM was trained on a massive dataset of 2 billion samples, whereas contrastive and MIM models can achieve competitive results with datasets that are 150 times smaller. This inefficiency poses a substantial barrier to their widespread adoption in resource-constrained environments. These limitations highlight the need for more efficient auto-regressive models for visual tasks. 

In this work, we propose XTRA, an auto-regressive vision model that leverages a Block Causal Mask to enhance sample and parameter efficiency. By employing Block Causal Masking, XTRA more effectively utilizes its modeling capacity to capture low-frequency structures essential for object recognition, rather than focusing on high-frequency details. Empirical results demonstrate that this approach enables XTRA to learn abstract and semantically meaningful representations using less data and smaller model sizes.

The key contributions of this paper are:
\begin{itemize}
    \item \textbf{High sample efficiency.} Although trained on 152$\times$ fewer samples (13.1M vs. 2B), XTRA ViT-H/14 outperforms the previous state-of-the-art auto-regressive model of the same size in top-1 average accuracy across 15 diverse image recognition benchmarks. 
    \item \textbf{High parameter efficiency.} XTRA ViT-B/16 outperforms auto-regressive models trained on ImageNet-1k in linear and attentive probing tasks, while using 7–16$\times$ fewer parameters (85M vs. 1.36B/0.63B).
\end{itemize}

%% file: sec/method_figure.tex
\begin{figure*}[ht]
    \centering
    \makebox[\textwidth]{ 
        \resizebox{0.9\textwidth}{!}{%
            \includegraphics{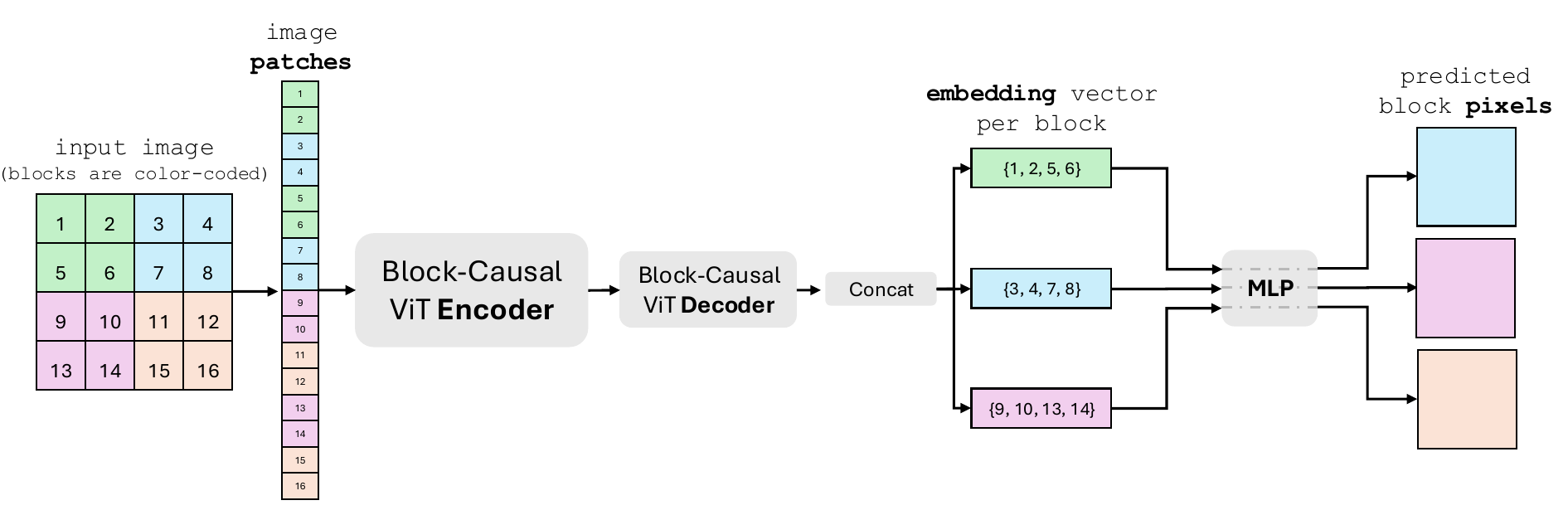}
        }
    }
    \caption{\textbf{XTRA Architecture.} Following ViT \cite{dosovitskiy2021an} an image is partitioned into a sequence of patches (numbered grid) and processed by a standard ViT encoder-decoder architecture with our proposed Block Causal Masking. I.e., causality is enforced at the block level with a rasterized pattern (see \Cref{fig:block_causal_masking} for detailed explanation). In the example above a block represents 2$\times$2 patches/tokens. The token representations within each block at the output of the decoder are concatenated in a predetermined order such that each block of pixels is represented by a single embedding vector. Finally, each block embedding vector is passed through an MLP (same MLP for all blocks) to predict \textbf{all} pixel values of the next block in the sequence.}
    \label{fig:arch}
\end{figure*}

%% file: sec/related_work.tex
\section{Related Work}
\subsection{Contrastive Self-Supervised Learning}
\label{section:contrastive_learning}
Contrastive Learning methods aim to maximize the similarity of two \textit{handcrafted} augmentations (also called views) of a given image, while preventing collapse. Collapse is defined as the trivial solution where all images in the dataset are assigned the same vector representation. The various methods differ by the way they prevent collapse. For instance, SimCLR \cite{chen2020simclr, chen2020big} tackles collapse by utilizing negative pairs, explicitly minimizing the similarity between different images. MoCo \cite{he2020momentum, chen2020improved}, BYOL \cite{grill2020bootstrap} and DINO \cite{caron2021emerging}, on the other hand, employ a momentum encoder (also utilized beyond contrastive methods) to mitigate collapse. SwAV \cite{caron2020unsupervised} takes a different route by relying on an external clustering algorithm to prevent collapse effectively. Self-Classifier \cite{amrani2022self}, meanwhile, counters collapse with the use of an explicit uniform prior. Lastly, SimSiam \cite{chen2021exploring} employs a stop-gradient operation on one of the views as its means to avert collapse. 

\subsection{Generative Self-Supervised Learning}
\label{section:generative_learning}
Generative self-supervised learning aims to learn representations that enable the prediction of masked regions within a sample based on the unmasked portions. While this paradigm has seen remarkable success in Natural Language Processing \cite{radford2019language,brown2020gpt3,kenton2019bert}, it has recently made its way into the domain of computer vision. Generative learning approaches in Computer Vision can be categorized into two categories, each with its own techniques and objectives. These two categories are further detailed in the following subsections.

\subsubsection{Masked Image Modeling}
\label{section:mim_methods}
Following the introduction of the Vision Transformer (ViT) by \cite{dosovitskiy2021an}, BEiT \cite{bao2022beit} extended the concept of masked language modeling from NLP to visual tasks via a Masked Image Modeling (MIM) approach, inspired by BERT \cite{kenton2019bert}. BEiT partitions an image into a grid of patches and tokenizes each patch into discrete visual tokens using latent codes obtained from a pre-trained discrete variational autoencoder \cite{rolfe2016discrete} (dVAE). The objective is to predict the masked visual tokens based on the unmasked patches, mirroring BERT's approach of recovering masked words using the surrounding text context.
iBOT \cite{zhou2021ibot} built on BEiT by replacing the fixed dVAE tokenizer with an online tokenizer, learned through a momentum encoder that is updated during training. Furthermore, iBOT integrates contrastive learning, minimizing the similarity between two augmented views of the same image to improve performance.
Other BERT-inspired methods, such as MAE \cite{he2022masked} and SimMIM \cite{xie2022simmim}, deviate from BEiT by abandoning discrete tokenization. Instead, they focus on directly predicting the pixel values of masked patches. MAE, in particular, adopts an asymmetric encoder-decoder architecture, where the encoder processes only visible patches, and a lightweight decoder reconstructs the missing patches from the latent representation and mask tokens. This approach has proven highly effective, demonstrating that pixel-level reconstruction can achieve impressive results in self-supervised visual learning.
Further MIM approaches, such as data2vec \cite{baevski2022data2vec,baevski2023efficient} and I-JEPA \cite{assran2023self}, move away from pixel reconstruction loss and instead predict the \textit{latent representations} of masked regions based on unmasked parts. These methods rely on a momentum encoder to avoid collapse—preventing all patches from being assigned the same representation. Lastly, Context Autoencoder \cite{chen2024context} (CAE) blends two pre-training tasks: (1) masked patch reconstruction (as seen in MAE and SimMIM) and (2) masked representation prediction (as seen in I-JEPA and data2vec).

\subsubsection{Auto-Regressive Image Modeling}
\label{section:aim_methods}
One of the early auto-regressive models for computer vision is iGPT \cite{chen2020generative} (Image-GPT). It is a sequence Transformer designed for auto-regressively predicting pixel values in images by minimizing the negative log-likelihood over a fixed set of pixel values. It represents a straightforward yet effective extension of GPT \cite{radford2019language} to image pixels. Although it did not surpass the performance of state-of-the-art Contrastive Learning methods, the main contribution of iGPT was demonstrating that the auto-regressive paradigm can be smoothly extended from language to vision. Primary limitations of iGPT, however, lie in its memory and computation requirements. First, due the quadratic nature of self-attention, processing a sequence of \textit{pixels} in an image is both memory and computationally expensive. Second, the sample- and parameter-efficiency of iGPT is considerably low in comparison to other self-supervised methods, as it requires more than 15$\times$ more parameters and more than 78$\times$ more samples for results to be competitive with self-supervised benchmarks on ImageNet. A more recent work is AIM \cite{el2024scalable} (Auto-regressive Image Models). AIM applies the standard auto-regressive objective using a causal Vision Transformer to a sequence of non-overlapping image patches (where a patch is embedded linearly following ViT), and predicts the pixel values of the next patch by minimizing the Mean Squared Error loss. The main contribution of AIM was showing that auto-regressive \textit{image} models exhibit similar scaling properties to auto-regressive \textit{language} models. I.e., the performance of the learned visual features scale with both the model capacity and the quantity of data, and the value of the objective function correlates with the performance of the model on downstream tasks. Yet, similarly to iGPT, the sample- and parameter-efficiency, though improved, is still low. Specifically, when trained on ImageNet-1k only, AIM requires more than 7$\times$ more parameters in comparison to other Contrastive and Masked Image Modeling methods for competitive results (see \cref{tab:aim_comparison_imagenet}). Additionally, for same model size (e.g., ViT-H/14), AIM requires more than 150$\times$ more samples for achieving stronger results on a diverse set of 15 image recognition benchmarks (see \cref{tab:downstream_perf}).

Building upon the foundation set by iGPT and AIM, our approach introduces an auto-regressive image model that leverages a Vision Transformer architecture with a novel Block Causal Mask (see \cref{section:method}). By processing blocks of image patches rather than individual patches or pixels, XTRA makes more efficient use of its modeling capacity to capture low-frequency structures that enhance object recognizability, rather than focusing on redundant high-frequency details. Empirically, this design allows us to substantially enhance sample- and parameter-efficiency (\cref{tab:downstream_perf} and \cref{tab:aim_comparison_imagenet}) while maintaining the simplicity and scalability that characterize auto-regressive models. 

%% file: sec/method.tex
\section{Method}
\label{section:method}

\begin{figure}[tb]
    \centering
    \makebox[\linewidth]{
        \resizebox{0.85\linewidth}{!}{%
            \includegraphics{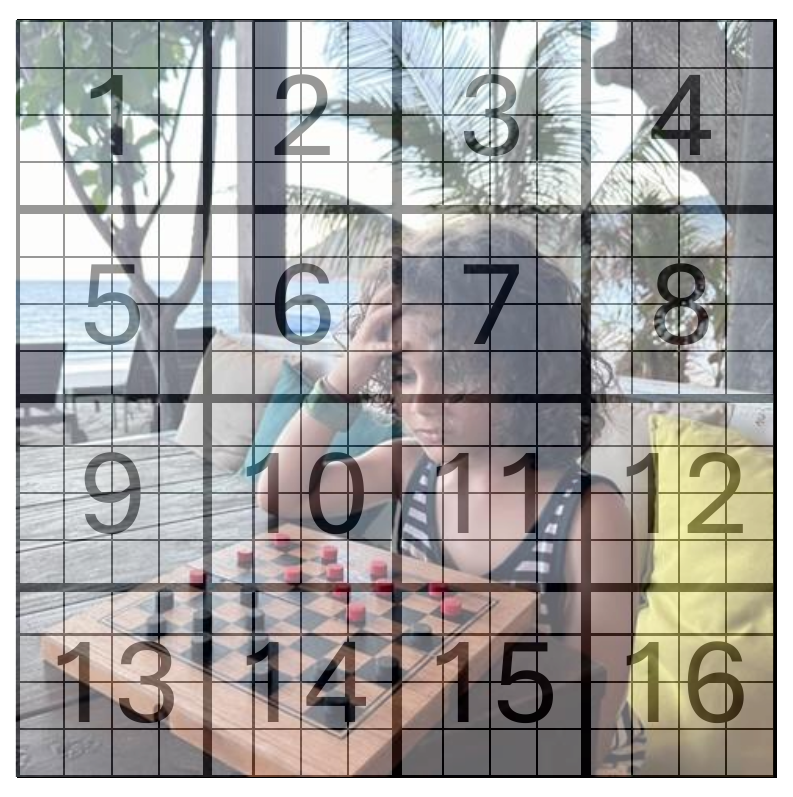}
        }
    }
    \caption{\textbf{Block Causal Masking.} In the image above the fine-grained grid represents a grid of patches (following ViT \cite{dosovitskiy2021an}) that are processed by the model. The coarse-grained (numbered) grid represents a grid of blocks, where each block represents 4$\times$4 patches/tokens. Block Causal Masking enforces causality at the block level with a rasterized pattern (numbered sequence), ensuring that tokens can attend to others within the same block and also to preceding blocks.}
    \label{fig:block_causal_masking}
\end{figure}
This section provides details on the XTRA architecture (\Cref{fig:arch}), including its use of \textit{Block Causal Masking} (\Cref{fig:block_causal_masking}), the training objective and loss function.
\paragraph{Architecture.} XTRA is an encoder-decoder network, where a Vision Transformer \cite{dosovitskiy2021an} (ViT) with \textit{Block Causal Masking} is used for both components. 

\paragraph{Block Causal Masking.} Block Causal Masking, visualized in \Cref{fig:block_causal_masking}, structures attention such that the image is divided into spatial blocks of $k \times k$ tokens, with causality enforced at the block level with a rasterized pattern. This structure allows for: (1) auto-regressive modeling with image regions bigger than a single patch size ; and (2) efficient modeling of both local (within-block) and global (cross-block) dependencies, ensuring that tokens can attend to others within the same block and also to preceding blocks.

\paragraph{Next Block Reconstruction.} At the output of the decoder, the token representations within each block are concatenated in a predetermined order (e.g., raster order) and passed through a fully-connected MLP to reconstruct the pixel values of the next block. This block-wise reconstruction strategy leverages the structured attention, enabling accurate pixel prediction for subsequent blocks based on previously processed regions.

\paragraph{Training Objective \& Loss.} The training objective employs a standard auto-regressive approach, generating predictions sequentially, where each block's prediction is based solely on previously observed blocks. The loss function used is the Mean Squared Error (MSE) applied to pixel values normalized per block, as per \cite{he2022masked,el2024scalable}:

\begin{equation}
\ell(\theta) = \frac{1}{N(K-1)}\sum_{n=1}^N\sum_{k=2}^K\|\hat{x}_k^n(\theta;x_{<k}^n)-x^n_k\|_2^2,
\end{equation}

where, $\theta$ represents the network's parameters, $N$ is the batch size, $K$ denotes the number of blocks in an image, $x^n_k$ is the ground truth pixel values for the $k$-th block in the $n$-th image, and $\hat{x}_k^n(\theta;x_{<k}^n)$ represents the reconstructed values based on the network parameters ($\theta$) and the preceding blocks in the sequence for the same image ($x_{<k}^n$).

%% file: sec/implementation_details.tex
\section{Implementation Details}
\label{section:implementation_details}
Default setting for the pre-training stage is in \Cref{tab:pt_hyperparameters}. Default setting for the attentive probing stage is in \Cref{tab:probe_hyperparameters}. Our training hyper-parameters are taken from AIM \cite{el2024scalable}.

\begin{table}
    \begin{center}
        \tiny
        \centering
        \setlength{\tabcolsep}{10pt}
        \resizebox{\linewidth}{!}{
        \begin{tabular}{l|cc}
            config & ViT-B/16 & ViT-H/14 \\
            \toprule
            Optimizer & \multicolumn{2}{c}{AdamW} \\
            Optimizer Momentum & \multicolumn{2}{c}{$\beta_1=0.9,\beta_2=0.95$} \\
            Peak learning rate & \multicolumn{2}{c}{$1e^{-3}$}\\
            Minimum Learning rate & \multicolumn{2}{c}{$1e^{-6}$} \\
            Weight decay & \multicolumn{2}{c}{0.05} \\
            Batch size & \multicolumn{2}{c}{2048} \\
            Patch size & 16$\times$16 px & 14$\times$14 px \\
            Block size & 64$\times$64 px & 56$\times$56 px \\
            Decoder width & 768 & 640 \\
            Decoder depth & \multicolumn{2}{c}{8} \\
            Gradient clipping & \multicolumn{2}{c}{1.0} \\
            Drop path rate & \multicolumn{2}{c}{0.2} \\
            Dataset & ImageNet-1K & ImageNet-21K \\
            Warmup epoch & 15 & 3 \\
            Total epochs & 800 & 100 \\
            Learning rate schedule & \multicolumn{2}{c}{cosine decay} \\
            Augmentations: \\
            \quad {\tt RandomResizedCrop} \\
            \qquad {\tt size} & 256px & 224px \\
            \qquad {\tt scale} & \multicolumn{2}{c}{[0.3, 1.0]} \\
            \qquad {\tt ratio} & \multicolumn{2}{c}{[0.75, 1.33]} \\
            \quad {\tt RandomHorizontalFlip} & \multicolumn{2}{c}{$p=0.5$} \\
        \end{tabular}}
    \end{center}
    \caption{\textbf{Pre-training hyperparameters.}}
    \label{tab:pt_hyperparameters}
    \end{table}

\begin{table}
    \begin{center}
        \tiny
        \centering
        \setlength{\tabcolsep}{4pt}
        \resizebox{\linewidth}{!}{
        \begin{tabular}{l|rr}
            config & IN-1k & Others \\
            \toprule
            Optimizer & \multicolumn{2}{c}{AdamW} \\
            Optimizer Momentum & \multicolumn{2}{c}{$\beta_1=0.9,\beta_2=0.999$} \\
            Peak learning rate grid & \multicolumn{2}{c}{[1, 3, 5, 10, 15, 20, 40, 80]~$\times1e^{-4}$} \\
            Minimum Learning rate & \multicolumn{2}{c}{$1e^{-5}$} \\
            Weight decay & \multicolumn{2}{c}{0.1} \\
            Batch size & 4k & [128, 256, 512]$^*$ \\
            Gradient clipping & \multicolumn{2}{c}{3.0} \\
            Drop path rate & \multicolumn{2}{c}{0.0} \\
            Warmup epochs & 10 & 0 \\
            Epochs & \multicolumn{2}{c}{100}  \\
            Learning rate schedule & \multicolumn{2}{c}{cosine decay} \\
            Augmentations: \\
            \quad {\tt RandomResizedCrop} \\
            \qquad {\tt size} & \multicolumn{2}{c}{224px} \\
            \qquad {\tt scale} & \multicolumn{2}{c}{[0.08, 1.0]} \\
            \qquad {\tt ratio} & \multicolumn{2}{c}{[0.75, 1.33]} \\
            \quad {\tt RandomHorizontalFlip} & \multicolumn{2}{c}{$p=0.5$} \\
            \quad {\tt AutoAugment} & \multicolumn{2}{c}{\texttt{rand-m9-mstd0.5-inc1}} \\
        \end{tabular}}
    \end{center}
    \caption{\textbf{Attentive probe hyperparameters.} $*$: Small, medium and large datasets (excluding imagenet) used batch sizes of 128, 256 and 512, respectively.}
    \label{tab:probe_hyperparameters}
\end{table}

%% file: sec/quantitative_results.tex
\section{Quantitative Results}
\label{section:quantitative_results}
In this section, we evaluate the performance of XTRA in comparison to state-of-the-art methods on two probing tasks: linear (LIN) and attentive (ATT). The linear probing task assesses the quality of the frozen pre-trained features by training a simple linear classifier, while the attentive probing task evaluates the ability to learn a lightweight attention mechanism over these frozen representations. These tasks provide a comprehensive measure of how effectively the pre-trained representations can be leveraged for downstream image recognition tasks with minimal fine-tuning.

\subsection{Transfer Learning}
\Cref{tab:downstream_perf} summarizes the results of attentive probing across 15 diverse image recognition benchmarks. These benchmarks cover a wide range of domains, including fine-grained recognition, medical imaging, satellite imagery, natural environments, and infographic images. This diversity highlights XTRA's robustness and adaptability across varied visual tasks. For detailed hyperparameters see \Cref{tab:probe_hyperparameters}. Specific details of each benchmark dataset are in \Cref{tab:dataset_descriptions} in the Appendix.

XTRA-H (ViT-H/14, 632M parameters), pre-trained on ImageNet-21K (filtered to 13.1M samples due to broken URL links), achieves superior or competitive performance on 9 out of 15 datasets, setting a new benchmark for generative models with the highest average accuracy across tasks. Notably, XTRA outperforms the previous state-of-the-art auto-regressive model, AIM-0.6B \cite{el2024scalable}, by 1.7\% when compared to AIM-0.6B trained on DFN-2B and by 0.6\% against AIM-0.6B trained on DFN-2B+ (80\% DFN-2B and 20\% ImageNet-1K). Despite training on 152$\times$ fewer samples (13.1M vs. 2B), XTRA’s sample efficiency enables it to learn rich, semantically meaningful representations, which lead to these improvements.

\begin{table*}[htb]
    \centering
    \setlength{\tabcolsep}{2pt}
    \renewcommand{\arraystretch}{1.2}
    \resizebox{1.0\linewidth}{!}{
    \begin{tabular}{llcccccccccccccccccc}
        \textbf{Model} & Arch. & Data & Cost$^\dagger$ & \small{\rotatebox{90}{IN-1k}} & \small{\rotatebox{90}{iNAT-18}} & \small{\rotatebox{90}{Cifar10}}  & \small{\rotatebox{90}{Cifar100}} & \small{\rotatebox{90}{Food101}} & \small{\rotatebox{90}{DTD}} & \small{\rotatebox{90}{Pets}} & \small{\rotatebox{90}{Cars}} & \small{\rotatebox{90}{iWildCam}} & \small{\rotatebox{90}{Camelyon17}} & \small{\rotatebox{90}{PCAM}} & \small{\rotatebox{90}{RxRX1}}  & \small{\rotatebox{90}{EuroSAT}} & \small{\rotatebox{90}{fMoW}} & \small{\rotatebox{90}{Infographic}} & Avg\\
         \toprule 
        \multicolumn{19}{l}{\textit{methods using extra view data augmentations}}
          \\
         DINO \cite{caron2021emerging} & ViT-B/8 & IN-1k & 19.2 & 80.1 & 66.0 & 97.8 & 87.3 & 89.5 & 78.4 & 92.3 & 89.2  & 58.5 & 93.7 & 90.2 & 6.1 & 98.2 & 57.0 & 41.1 & 75.0 \\
         iBOT \cite{zhou2021ibot} & ViT-L/16 & IN-21k & 1.4 & 83.5 & 70.5 & 99.2 & 93.3 & 93.5 & 81.6 & 92.8 & 90.8   & 61.8 & 94.5 & 90.0 & 5.9 & 98.0 & 60.3 & 47.7 & 77.6 \\
         \midrule
         \midrule
         \multicolumn{19}{l}{\textit{methods without view data augmentations}} \\
         \hdashline
         \multicolumn{19}{l}{\small{\textit{--- masked image modeling methods}}} \\
         BEiT \cite{bao2022beit} & ViT-L/14 & IN-21k & \textbf{4.2} &
         62.2 & 44.4 & 94.4 & 78.7 & 79.0 & 64.0 & 80.9 & 69.5  & 52.0 &	92.8 & 88.2 & 4.2 &	97.5 &	47.7 & 25.9 & 65.4  \\
        MAE-H \cite{he2022masked} & ViT-H/14 & IN-1k & 8.0
        & \textbf{80.9} %
        & 64.6 %
        & 97.1
        & 85.8
        & 90.2 %
        & 78.1 %
        & \textbf{95.0} %
        & \textbf{93.7} %
        & 58.1    %
        & 94.2    %
        & 89.8    %
        & 5.4   %
        & 98.1 %
        & 56.9 %
        & 42.2
        & 75.3
        \\
        \hdashline
         \multicolumn{19}{l}{\small{\textit{--- auto-regressive image modeling methods}}} \\
         AIM-0.6B \cite{el2024scalable}  & ViT-H/14 & DFN-2B & 20.7 
          & -- %
          & -- %
          & -- %
          & -- %
          & -- %
          & -- %
          & -- %
          & -- %
          & -- %
          & -- %
          & -- %
          & -- %
          & -- %
          & -- %
          & -- %
          & 74.5 \\
         AIM-0.6B \cite{el2024scalable} & ViT-H/14 & DFN-2B+ & 20.7 
          & 78.5 %
          & 64.0 %
          & 97.2
          & 86.8
          & 90.1 %
          & \textbf{80.1} %
          & 93.0 %
          & 93.0 %
          & 57.9 %
          & 94.3  %
          & \textbf{90.0} %
          & \textbf{7.8} %
          & 98.4 %
          & 58.3 %
          & \textbf{45.2}
          & 75.6
          \\
          \rowcolor{lightgreen} XTRA-H (ours) & ViT-H/14 & IN-21k & 5.8 &
            \textbf{80.9}
          & \textbf{67.0}
          & \textbf{98.2}
          & \textbf{90.0}
          & \textbf{90.8}
          & 79.7
          & 93.7
          & 93.1
          & \textbf{59.5}
          & 93.3 
          & \textbf{90.0}
          & 5.7 
          & \textbf{98.5}
          & \textbf{58.6}
          & 43.9 
          & \textbf{76.2} \\
          \bottomrule
    \end{tabular}}
    \caption{\textbf{Downstream evaluation with a frozen trunk.} Similarly to AIM, We assess the quality of XTRA by evaluating against a diverse set of 15 image recognition benchmarks (specific details of each dataset are in \Cref{tab:dataset_descriptions} in the Appendix). XTRA and the baseline methods are evaluated using attentive probing with a frozen trunk. The attentive probing results for all other methods are from AIM \cite{el2024scalable}. ${\dagger}$: Computational cost is estimated using $Parameters \times Samples\times Epochs \times Views^2 \times Tokens^2$ (see \Cref{appendix:cost_explanation} in Appendix for detailed explanation of this formula). }
    \label{tab:downstream_perf}
\end{table*}

\subsection{ImageNet-1K}
Here, we focus on models trained and evaluated exclusively on ImageNet-1K. 

In \Cref{tab:aim_comparison_imagenet}, we compare XTRA-B (ViT-B/16, 85M parameters) to prior auto-regressive image models. Despite having 16$\times$ fewer parameters, XTRA-B achieves a 5.0\% higher accuracy in the linear probing task compared to iGPT-L (1.36B parameters). In the attentive probing task, XTRA-B outperforms AIM-0.6B (0.63B parameters) by 3.3\%, using 7.5$\times$ fewer parameters. These results demonstrate XTRA’s high parameter efficiency, delivering superior performance with a fraction of the model size, which translates to faster inference times and lower computational costs.

Next, we compare XTRA-B with other state-of-the-art models that utilize the same ViT-B/16 backbone (\Cref{tab:frozen_trunk_eval_imagenet}). To ensure fair comparisons, we limit the analysis to models trained with single-crop views and without image-specific augmentations or multi-crop training. Under these conditions, XTRA-B establishes a new state-of-the-art in both linear and attentive probing tasks on ImageNet-1K. Consistent improvements are observed across training durations of 300 and 800 epochs. Notably, XTRA-B trained for 800 epochs surpasses both data2vec and MAE, which require 1600 epochs to train, further highlighting XTRA’s ability to achieve competitive results with significantly fewer training resources.

These findings underscore the advantages of the auto-regressive pre-training objective used by XTRA, particularly in terms of sample and parameter efficiency. XTRA achieves competitive results with reduced training time and computational resources, making it a highly efficient approach compared to previous methods.

\begin{table}
  \centering
  \setlength{\tabcolsep}{8pt}
  \begin{tabular}{lccc}
    \textbf{Method} & \textbf{\# Params (M)} & \textbf{LIN} & \textbf{ATT} \\
    \toprule
    iGPT-L$^\dagger$ \cite{chen2020generative} & 1362 & 65.2 & - \\
    AIM-0.6B \cite{el2024scalable} & 600 & - & 73.5 \\
    \rowcolor{lightgreen} XTRA (ours) & 85 & \textbf{70.2} & \textbf{76.8} \\
    \bottomrule
  \end{tabular}
  \caption{\textbf{Comparison to previous Auto-Regressive Image Modeling methods}. LIN: linear probing accuracy. ATT: attentive probing accuracy. All models were pre-trained and evaluated with ImageNet-1k. $\dagger$: iGPT use linear probing by concatenating the output of 5 layers, which indirectly inflates the capacity of the evaluation head. 
  }
  \label{tab:aim_comparison_imagenet}
\end{table}

\begin{table}
  \centering
  \setlength{\tabcolsep}{10pt}
  \begin{tabular}{lccc}
    \textbf{Method} & \textbf{Epochs} & \textbf{LIN} & \textbf{ATT} \\
    \toprule
    \multicolumn{4}{l}{\textit{methods using extra augmented views$^{\diamond}$}} \\
    MoCo-v3 \cite{chen2021empirical} & 300 & 76.2 & 77.0 \\
    DINO \cite{caron2021emerging} & 400 & 77.3 & 77.8 \\
    iBOT \cite{zhou2021ibot} & 400 & 79.5 & 79.8 \\
    \midrule
    \multicolumn{4}{l}{\textit{methods using extra masked views$^{\star}$}} \\
    I-JEPA \cite{assran2023self} & 600 & 70.9 & - \\
    StoP \cite{barstochastic} &  600 & 72.6 & - \\
    \midrule
    \midrule
    \multicolumn{4}{l}{\textit{methods with a single view}} \\
    MAE \cite{he2022masked} & 300 & 61.5 & 71.1 \\
    CAE$^{\dagger}$ \cite{chen2024context} & 300 & 64.1 & 73.8 \\ 
    \rowcolor{lightgreen} XTRA (ours) & 300 & \textbf{66.1} & \textbf{74.3} \\
    \hdashline
    SimMIM \cite{xie2022simmim} & 800 & 56.7 & - \\ 
    MAE \cite{he2022masked} & 1600 & 67.8 & 74.2 \\ 
    Data2Vec \cite{baevski2022data2vec} & 1600 & 68.0 & - \\
    CAE$^{\dagger}$ \cite{chen2024context} & 800 & 68.6 & 75.9 \\ 
    \rowcolor{lightgreen} XTRA (ours) & 800 & \textbf{70.2} & \textbf{76.8} \\
    \bottomrule
  \end{tabular}
  \caption{\textbf{ViT-B ImageNet-1k evaluation with a frozen trunk.} All models were trained and evaluated exclusively on ImageNet-1K. LIN: linear probing accuracy. ATT: attentive probing accuracy. 
  $\star$: I-JEPA and StoP use multi-mask views. For each 1 context block mask, 4 target blocks masks are sampled; both results are taken from StoP \cite{barstochastic} for linear probing using the single last layer.
  $\diamond$: MoCo-v3, DINO and iBOT use multi-crop \textit{hand-crafted} view augmentations. MoCo-v3: 2 crops. DINO and iBOT: 12 crops. Thus, the number of effective epochs for both $\star$ and $\diamond$ is larger (equivalent to taking a larger number of epochs compared to one-crop augmentation).
  In contrast, Data2Vec, MAE, CAE and XTRA use a single view without any image-specific augmentations (i.e., only random crops). 
  The attentive probing results for all other methods are from CAE \cite{chen2024context}. ${\dagger}$: denotes using the DALL-E tokenizer (trained with d-VAE on 400M images).}
  \label{tab:frozen_trunk_eval_imagenet}
\end{table}

%% file: sec/ablations.tex
\section{Ablation Study}
In this section, we evaluate the impact of the main components of XTRA. For each experiment, we pre-train a ViT-B/16 model for 100 epochs, followed by a supervised attentive probing training stage using the ImageNet-1K training set. We report the attentive probing accuracy on the ImageNet-1K validation set in \Cref{tab:ablation}. 

\paragraph{Block size \& multi-block prediction.} Block size is a critical component of XTRA. \Cref{tab:ablation_block_size} shows results for various block sizes: 16$\times$16, 32$\times$32, and 64$\times$64 pixels, corresponding to a single token/patch, 2$\times$2 tokens, and 4$\times$4 tokens, respectively. We also experimented with `multi-block' prediction, where multiple blocks are predicted based on prior context (e.g., predicting the next two blocks). As shown, this does not significantly impact the results, except in the case of a single token versus two tokens (as seen in the first row of the table). 
When the block size is set to a single patch/token (16$\times$16 pixels), it replicates the standard auto-regressive image model (AIM \cite{el2024scalable}), with \textit{all} other training details kept identical. Notably, increasing the block size has a substantial effect on performance, with a 3.0\% accuracy improvement when comparing a block size of 16$\times$16 pixels (AIM re-implementation) to 64$\times$64 pixels (XTRA). This demonstrates the clear benefits of XTRA and our proposed Block Causal Masking in auto-regressive image models.

\paragraph{Impact of block size relative to resolution.} To further explore the impact of block size, we evaluate various block sizes in relation to the image resolution, as shown in \Cref{tab:ablation_block_to_res_ratio}. Specifically, we compare two different resolutions (224 and 256 pixels) with corresponding patch sizes of 14 and 16 pixels, respectively. We observe that when the block size to resolution ratio is kept constant across experiments (as reflected in each row of the table), the results remain consistent, particularly when the block size to resolution is sufficiently large (e.g., 4/256 or 16/256). This consistency holds even when the patch size, image size, and absolute block size (in pixels) differ, suggesting that for auto-regressive image models, the block size to resolution ratio is the key factor determining performance, rather than the absolute block size.

\paragraph{Loss function.} In \Cref{tab:ablation_loss}, we compare L1 loss (Mean Absolute Error) with L2 loss (Mean Squared Error). The results show that L2 loss outperforms L1 loss (+1.0\%).

\paragraph{Auto-regressive pattern.} In \Cref{tab:ablation_ar_pattern}, we compare the simple raster pattern with a fixed random pattern. The results indicate that the raster pattern significantly outperforms the random pattern, with an improvement of 11.9\%. We hypothesize that this large margin arises from the substantial difficulty introduced by randomizing the sequence of blocks in the auto-regressive prediction task. For example, certain random permutations may force the model to predict the top-left block of pixels based \textit{solely} on the bottom-right block, an arrangement which, for non-object-centric images, may yield an unsolvable task and limit the model’s ability to learn semantically meaningful representations. By contrast, predicting the next nearest block in raster order is generally feasible, even with only a single prior block as context, allowing for more coherent learning.

\paragraph{Decoder depth \& width.} Unlike previous auto-regressive image models (AIM and iGPT), which use only an encoder, XTRA is an encoder-decoder architecture. Although the decoder is not crucial to XTRA’s success, it serves two purposes: (1) as observed in previous works (e.g., AIM and MAE), the final encoder layers often specialize in pixel reconstruction rather than semantic recognition, so adding a decoder can localize this reconstruction specialization, allowing the encoder output to retain more abstract representations; and (2) it enables down-sampling of the embedding size before pixel predictions, improving reconstruction efficiency. In \Cref{tab:ablation_decoder_depth} and \Cref{tab:ablation_decoder_width}, we assess the impact of the decoder’s depth and width.
The results show that when the width is fixed at 384 (half the encoder's width), the decoder depth has little effect on performance. However, with a fixed depth, increasing the width to 768 (matching the encoder’s width) yields a substantial 2.0\% performance improvement. Notably: (1) as with previous methods, the decoder is used only during pre-training for block reconstruction, ensuring that linear/attentive probing comparisons with other methods remain fair; and (2) in \Cref{tab:ablation_block_size}, we also re-implement AIM with an encoder-decoder structure, validating that XTRA’s Block Causal Masking provides performance gains irrespective of architecture. 

\begin{table*}
    \centering
    \small
    \begin{tabular}{ccc}       
        \begin{subtable}[t]{0.3\textwidth}
            \centering
            \begin{tabular}{c|cc}
             & \multicolumn{2}{c}{\# of blocks to predict} \\
            block size & 1 & 2 \\
            \toprule
            16$\times$16 & 64.6 & 65.1 \\
            32$\times$32 & 67.4 & 67.3 \\
            64$\times$64 & \cellcolor{lightgray} \textbf{67.6} & 67.4 \\
            \end{tabular}
            \caption{\textbf{Block size (pixels) \& multi-block prediction.}}
            \label{tab:ablation_block_size}
        \end{subtable} & 

        \begin{subtable}[t]{0.3\textwidth}
            \centering
            \begin{tabular}{c|cc}
             & \multicolumn{2}{c}{res.$\vert$patch size} \\
            block / res. ratio & 256$\vert$16 & 224$\vert$14 \\
            \toprule
            1/256 & 64.6 & 65.2 \\
            4/256 & 67.4 & 67.3 \\
            16/256 & \cellcolor{lightgray} 67.6 & \textbf{67.7 }
            \end{tabular}
            \caption{\textbf{Block size relative to resolution.}}
            \label{tab:ablation_block_to_res_ratio}
        \end{subtable} &
        
        \begin{subtable}[t]{0.3\textwidth}
            \centering
            \begin{tabular}{c|c}
                Function & Accuracy \\
                \toprule
                L1 (MAE) & 66.6 \\
                L2 (MSE) & \cellcolor{lightgray} \textbf{67.6} \\\\\\
            \end{tabular}
            \caption{\textbf{Loss function.}}
            \label{tab:ablation_loss}
        \end{subtable} \\

        \begin{subtable}[t]{0.3\textwidth}
            \centering
            \begin{tabular}{c|c}
                AR Pattern & Accuracy \\
                \toprule
                Fixed Random & 55.7 \\
                Raster & \cellcolor{lightgray} \textbf{67.6} \\\\\\\\
            \end{tabular}
            \caption{\textbf{Auto-regressive pattern.}}
            \label{tab:ablation_ar_pattern}
        \end{subtable} &
        
        \begin{subtable}[t]{0.3\textwidth}
            \centering
            \begin{tabular}{c|c}
                blocks & Accuracy \\
                \toprule
                1 & 67.4 \\
                2 & \textbf{67.9} \\
                4 & 67.6 \\
                8 & \cellcolor{lightgray} 67.6 \\
                16 & 67.8 \\
            \end{tabular}
            \caption{\textbf{Decoder depth.}}
            \label{tab:ablation_decoder_depth}
        \end{subtable} &
        
        \begin{subtable}[t]{0.3\textwidth}
            \centering
            \begin{tabular}{c|c}
                width & Accuracy \\
                \toprule
                192 & 67.1 \\
                384 & \cellcolor{lightgray} 67.6 \\
                576 & 67.8 \\
                768 & \textbf{69.6} \\ \\
            \end{tabular} 
            \caption{\textbf{Decoder width.}}
            \label{tab:ablation_decoder_width}
        \end{subtable}
    \end{tabular}
    \caption{\textbf{Ablation study.} Default settings for ablation baseline are marked in \colorbox{lightgray}{gray}. Best in \textbf{bold}.}
    \label{tab:ablation}
\end{table*}

%% file: sec/discussion.tex
\section{Discussion}
\noindent Our model employs \textit{Block Causal Masking} to improve both \textbf{sample efficiency} and \textbf{parameter efficiency} by structuring autoregressive attention to better align with the 2D structure of images.

\paragraph{Sample efficiency.} Standard autoregressive models process tokens sequentially, making it inefficient to capture local spatial structure. In contrast, our block causal mask groups tokens into local regions, allowing \textbf{unrestricted intra-block attention} while enforcing causality between blocks. This design improves sample efficiency by explicitly modeling local pixel dependencies, which reduces the number of training samples required to learn meaningful representations. Additionally, by allowing free attention within each block, local structures are captured in fewer autoregressive steps, making representation learning faster and requiring less data to generalize effectively.

\paragraph{Parameter efficiency.} The block-wise autoregressive factorization also enhances parameter efficiency by reducing the depth needed to model long-range dependencies. Since tokens within a block attend freely to each other, information propagates efficiently without requiring excessively deep hierarchies. This leads to a more compact and computationally efficient model, as local structures are captured without redundant layers or parameters.

\paragraph{Empirical Validation.} These theoretical advantages manifest in practical improvements. As shown in Section~\ref{section:quantitative_results}, models with block causal masking achieve \textbf{higher accuracy with fewer training samples} and perform competitively even with fewer parameters compared to fully autoregressive baselines. 
%These results confirm that designing autoregressive attention to align with the spatial structure of images improves both training efficiency and inference performance. This highlights the broader effectiveness of structured autoregressive modeling for vision tasks, where efficient information propagation is essential.

%% file: sec/qualitative_results.tex
\begin{figure*}[hbt]
    \centering
    \makebox[\textwidth]{ 
        \resizebox{\textwidth}{!}{%
            \includegraphics{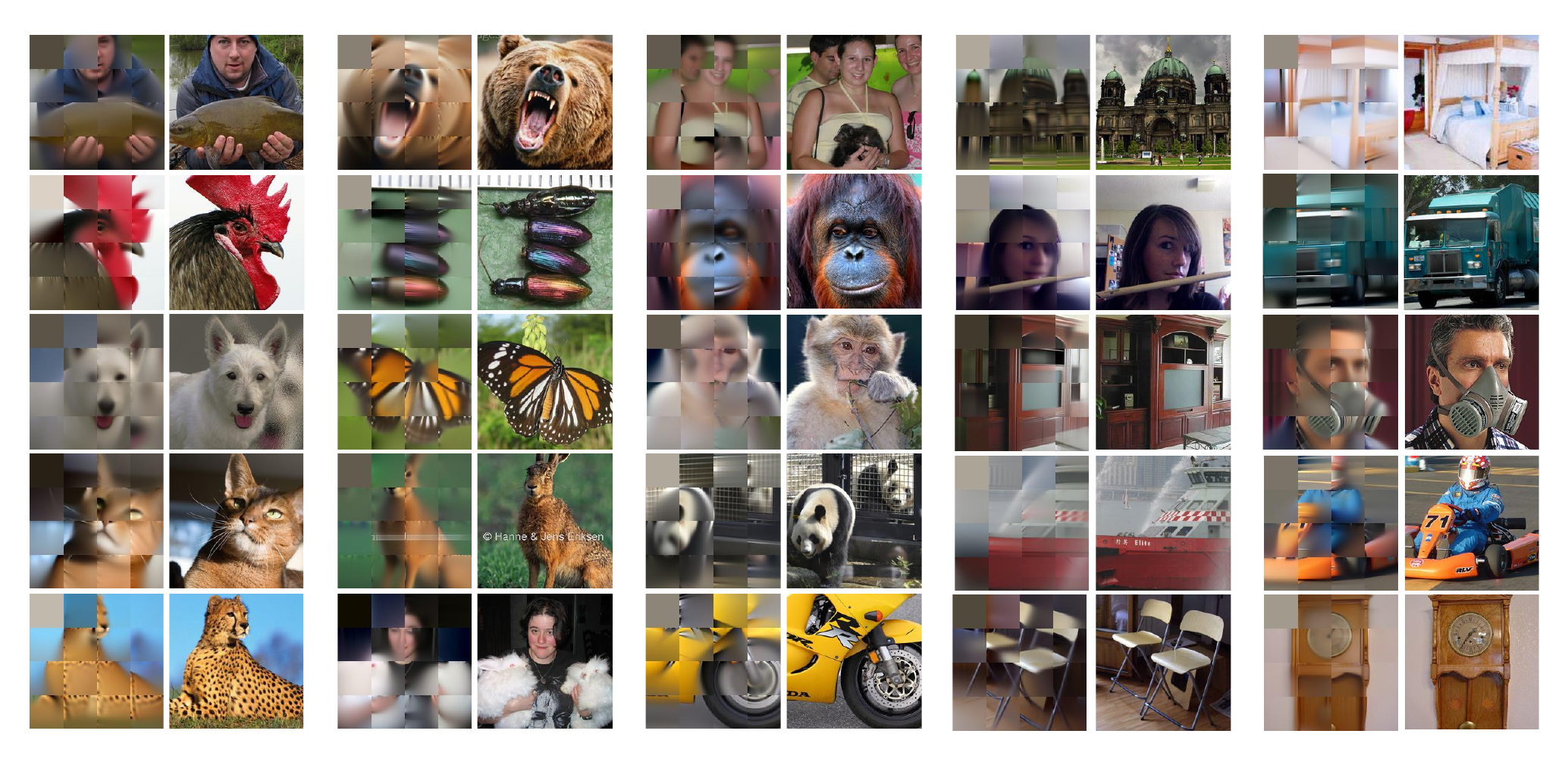}
        }
    }
    \caption{\textbf{Visualization of XTRA's Predictions on the ImageNet-1k Validation Set.} XTRA generates predictions auto-regressively, producing one block of pixels at a time, with each new block conditioned on the preceding sequence of ground-truth blocks. \textit{Note that no loss is applied to the left upper block (first in sequence), yet it is of different color from image to image due to post-generation per block de-normalization, since loss is applied to normalized block pixels.}}
    \label{fig:qualitative_results}
\end{figure*}

\section{Qualitative Results}
XTRA captures semantically meaningful representations by learning to predict unseen pixel values based on prior contextual information. In \Cref{fig:qualitative_results}, we present visualizations of the generative capabilities of XTRA to better understand the model's learned knowledge visually. Although XTRA's primary focus is on representation learning rather than generating photorealistic images, these visualizations offer valuable insights into the model's predictive reasoning. Notably, the effects of the Block Causal Mask are immediately observable: some blocks lack details of entire objects due to limited information from previous image blocks. Yet, even with these gaps, the predictions remain plausible given the current context.

%% file: sec/summary.tex
\section{Summary}
Recent advancements in Deep Learning have focused on simple, efficient, and scalable algorithms. Notably, the field of Natural Language Processing (NLP) has made remarkable progress by scaling architecture, data, and compute resources.
In this work, we aimed to create a similarly simple, efficient, and scalable approach for the field of Computer Vision, inspired by the successful scaling strategies applied to auto-regressive language models. We proposed XTRA, an auto-regressive image model that employs a novel Block Causal Masking technique. This model predicts the next block of pixels based on prior context, demonstrating significant improvements in sample and parameter efficiency compared to existing auto-regressive image models. Specifically, our empirical results indicate that XTRA achieves 152$\times$ greater sample efficiency and 7–16$\times$ improved parameter efficiency.
We hope our work inspires further exploration in scalable auto-regressive models for Computer Vision.

%% file: sec/suppl.tex
\clearpage
\setcounter{page}{1}
\maketitlesupplementary

\section{Benchmark Datasets}
\label{sec:benchmark_datasets}

\begin{table}[htb!]
    \centering
    \setlength{\tabcolsep}{8pt}
    \begin{tabular}{l|r|r|r}
        Dataset & train & test & classes \\
         \toprule
         Imagenet-1k~\cite{deng2009imagenet} & 1,281,167 & 50,000 & 1000 \\
         iNAT-18~\cite{inaturalist18}     & 437,513 & 24,426 & 8142 \\
         CIFAR-10~\cite{krizhevsky2009learning}   & 50,000 & 10,000 & 10 \\
         CIFAR-100~\cite{krizhevsky2009learning}  & 50,000 & 10,000 & 100 \\
         Food101~\cite{bossard2014food}     & 75,750 & 25,250 & 101 \\ 
         DTD~\cite{cimpoi2014describing}       & 3,760 & 1,880 & 47 \\
         Pets~\cite{parkhi2012cats}      & 3,680 & 3,669 & 37 \\
         Cars~\cite{krause20133d}  & 8,144 & 8,041 & 196 \\
         iWildCam~\cite{beery2021iwildcam}    & 129,809 & 14961 & 182 \\
         Camelyon17~\cite{bandi2018detection}  & 302,436 & 34904 & 2 \\
         PCAM~\cite{veeling2018rotation}      & 262,144 & 32768 & 2 \\
         RxRx1~\cite{taylor2019rxrx1}      & 40,612 & 9854 & 1139 \\
         EuroSAT~\cite{helber2019eurosat}     & 16,200 & 5400 & 10 \\
         fMoW~\cite{christie2018functional}       & 76,863 & 19915 & 62 \\
         Infograph~\cite{peng2019moment} & 36,023 & 15,582 & 345 \\
    \end{tabular}
    \caption{\textbf{Evaluation benchmarks.} We provide the references, the number of images in the train and test sets, and the number of categories of all the 15 recognition benchmarks used in this work. Table taken from \cite{el2024scalable}.}
    \label{tab:dataset_descriptions}
\end{table}

\section{Computational Cost Estimation}
\label{appendix:cost_explanation}
In \Cref{tab:computational_cost}, we \textit{estimate} the computational cost of each method using the following simplified formula:

\begin{equation}
    \text{Cost} = \text{Parameters} \times \text{Samples} \times \text{Epochs} \times \text{Views}^2 \times \text{Tokens}^2
\end{equation}

This formula provides an approximate scaling relationship rather than an exact measurement, as it does not account for hardware optimizations, model-specific efficiencies, or parallelization effects.

\begin{itemize}
    \item \textbf{Parameters (Linear):} The number of parameters in the model determines the size of weight matrices involved in computation. Since most architectures perform matrix multiplications that scale with the number of parameters, computation cost is approximately proportional to this term.
    \item \textbf{Samples (Linear):} The number of training samples contributes linearly since each sample requires a forward and backward pass.
    \item \textbf{Epochs (Linear):} The number of epochs scales cost linearly because training for more epochs means repeating the entire dataset multiple times.
    \item \textbf{Views (Squared):} If a method processes multiple views of the same data (e.g., contrastive learning with augmentations), the computational cost increases quadratically. This is because each pairwise interaction between views often involves computing similarities or attention across all view combinations.
    \item \textbf{Tokens (Squared):} The number of tokens per sample affects cost quadratically because self-attention mechanisms in transformers require $\mathcal{O}(\text{Tokens}^2)$ operations per forward pass.
\end{itemize}

For clarity, the numbers reported in \Cref{tab:downstream_perf} are divided by $10^{22}$, as the absolute units of computation do not impact the relative comparisons between methods.

While this formula captures key scaling behaviors, it does not precisely reflect real-world training cost due to factors like activation memory usage, hardware acceleration, and mixed precision training. Nonetheless, it serves as a useful proxy for comparing methods at scale.

Compared to MAE, AIM, and DINO, XTRA demonstrates greater effectiveness. While BEiT is slightly more efficient, XTRA outperforms it by +10.8\% in accuracy. iBOT, which integrates contrastive learning with masked image modeling, achieves better accuracy at a lower computational cost.

\begin{table}[t]
    \centering
    \small
    \begin{tabular}{lccccccc} 
    Name & \small{\rotatebox{90}{Parameters}} & \small{\rotatebox{90}{Samples}} & \small{\rotatebox{90}{Epochs}} & \small{\rotatebox{90}{Views}} & \small{\rotatebox{90}{Tokens}} & Cost & Acc. \\
    \midrule
    DINO & 85M & 1.2M & 800 & 2 & 768 & 19.2e22 & 75.0 \\
    iBOT & 307M & 1.2M & 250 & 2 & 196 & 1.4e22 & 77.6 \\
    \midrule
    BEiT & 307M & 14M & 150 & 1 & 256 & 4.2e22 & 65.4 \\
    MAE & 632M & 1.2M & 1600 & 1 & 256 & 8.0e22 & 75.3 \\
    AIM & 632M & 2B & 2.5 & 1 & 256 & 20.7e22 & 75.6\\
    XTRA & 632M & 14M & 100 & 1 & 256 & 5.8e22 & 76.2 \\
    \end{tabular}
    \caption{\textbf{Computational cost comparison.}}
    \label{tab:computational_cost}
\end{table}

%% file: main.bbl
\begin{thebibliography}{50}
\providecommand{\natexlab}[1]{#1}
\providecommand{\url}[1]{\texttt{#1}}
\expandafter\ifx\csname urlstyle\endcsname\relax
  \providecommand{\doi}[1]{doi: #1}\else
  \providecommand{\doi}{doi: \begingroup \urlstyle{rm}\Url}\fi

\bibitem[Amrani et~al.(2022)Amrani, Karlinsky, and Bronstein]{amrani2022self}
Elad Amrani, Leonid Karlinsky, and Alex Bronstein.
\newblock Self-supervised classification network.
\newblock In \emph{European Conference on Computer Vision}, pages 116--132.
  Springer, 2022.

\bibitem[Assran et~al.(2023)Assran, Duval, Misra, Bojanowski, Vincent, Rabbat,
  LeCun, and Ballas]{assran2023self}
Mahmoud Assran, Quentin Duval, Ishan Misra, Piotr Bojanowski, Pascal Vincent,
  Michael Rabbat, Yann LeCun, and Nicolas Ballas.
\newblock Self-supervised learning from images with a joint-embedding
  predictive architecture.
\newblock In \emph{Proceedings of the IEEE/CVF Conference on Computer Vision
  and Pattern Recognition}, pages 15619--15629, 2023.

\bibitem[Baevski et~al.(2022)Baevski, Hsu, Xu, Babu, Gu, and
  Auli]{baevski2022data2vec}
Alexei Baevski, Wei-Ning Hsu, Qiantong Xu, Arun Babu, Jiatao Gu, and Michael
  Auli.
\newblock Data2vec: A general framework for self-supervised learning in speech,
  vision and language.
\newblock In \emph{International Conference on Machine Learning}, pages
  1298--1312. PMLR, 2022.

\bibitem[Baevski et~al.(2023)Baevski, Babu, Hsu, and
  Auli]{baevski2023efficient}
Alexei Baevski, Arun Babu, Wei-Ning Hsu, and Michael Auli.
\newblock Efficient self-supervised learning with contextualized target
  representations for vision, speech and language.
\newblock In \emph{International Conference on Machine Learning}, pages
  1416--1429. PMLR, 2023.

\bibitem[Bandi et~al.(2018)Bandi, Geessink, Manson, Van~Dijk, Balkenhol,
  Hermsen, Bejnordi, Lee, Paeng, Zhong, et~al.]{bandi2018detection}
Peter Bandi, Oscar Geessink, Quirine Manson, Marcory Van~Dijk, Maschenka
  Balkenhol, Meyke Hermsen, Babak~Ehteshami Bejnordi, Byungjae Lee, Kyunghyun
  Paeng, Aoxiao Zhong, et~al.
\newblock From detection of individual metastases to classification of lymph
  node status at the patient level: the camelyon17 challenge.
\newblock \emph{IEEE transactions on medical imaging}, 38\penalty0
  (2):\penalty0 550--560, 2018.

\bibitem[Bao et~al.(2022)Bao, Dong, Piao, and Wei]{bao2022beit}
Hangbo Bao, Li Dong, Songhao Piao, and Furu Wei.
\newblock {BE}it: {BERT} pre-training of image transformers.
\newblock In \emph{International Conference on Learning Representations}, 2022.

\bibitem[Bar et~al.(2024)Bar, Bordes, Shocher, Assran, Vincent, Ballas,
  Darrell, Globerson, and LeCun]{barstochastic}
Amir Bar, Florian Bordes, Assaf Shocher, Mido Assran, Pascal Vincent, Nicolas
  Ballas, Trevor Darrell, Amir Globerson, and Yann LeCun.
\newblock Stochastic positional embeddings improve masked image modeling.
\newblock In \emph{International conference on machine learning}. PMLR, 2024.

\bibitem[Beery et~al.(2021)Beery, Agarwal, Cole, and
  Birodkar]{beery2021iwildcam}
Sara Beery, Arushi Agarwal, Elijah Cole, and Vighnesh Birodkar.
\newblock The iwildcam 2021 competition dataset.
\newblock \emph{arXiv preprint arXiv:2105.03494}, 2021.

\bibitem[Bossard et~al.(2014)Bossard, Guillaumin, and
  Van~Gool]{bossard2014food}
Lukas Bossard, Matthieu Guillaumin, and Luc Van~Gool.
\newblock Food-101--mining discriminative components with random forests.
\newblock In \emph{Computer vision--ECCV 2014: 13th European conference,
  zurich, Switzerland, September 6-12, 2014, proceedings, part VI 13}, pages
  446--461. Springer, 2014.

\bibitem[Brown et~al.(2020)Brown, Mann, Ryder, Subbiah, Kaplan, Dhariwal,
  Neelakantan, Shyam, Sastry, Askell, Agarwal, Herbert-Voss, Krueger, Henighan,
  Child, Ramesh, Ziegler, Wu, Winter, Hesse, Chen, Sigler, Litwin, Gray, Chess,
  Clark, Berner, McCandlish, Radford, Sutskever, and Amodei]{brown2020gpt3}
Tom Brown, Benjamin Mann, Nick Ryder, Melanie Subbiah, Jared~D Kaplan, Prafulla
  Dhariwal, Arvind Neelakantan, Pranav Shyam, Girish Sastry, Amanda Askell,
  Sandhini Agarwal, Ariel Herbert-Voss, Gretchen Krueger, Tom Henighan, Rewon
  Child, Aditya Ramesh, Daniel Ziegler, Jeffrey Wu, Clemens Winter, Chris
  Hesse, Mark Chen, Eric Sigler, Mateusz Litwin, Scott Gray, Benjamin Chess,
  Jack Clark, Christopher Berner, Sam McCandlish, Alec Radford, Ilya Sutskever,
  and Dario Amodei.
\newblock Language models are few-shot learners.
\newblock In \emph{Advances in Neural Information Processing Systems}, pages
  1877--1901. Curran Associates, Inc., 2020.

\bibitem[Caron et~al.(2020)Caron, Misra, Mairal, Goyal, Bojanowski, and
  Joulin]{caron2020unsupervised}
Mathilde Caron, Ishan Misra, Julien Mairal, Priya Goyal, Piotr Bojanowski, and
  Armand Joulin.
\newblock Unsupervised learning of visual features by contrasting cluster
  assignments.
\newblock In \emph{Advances in Neural Information Processing Systems}, 2020.

\bibitem[Caron et~al.(2021)Caron, Touvron, Misra, J\'egou, Mairal, Bojanowski,
  and Joulin]{caron2021emerging}
Mathilde Caron, Hugo Touvron, Ishan Misra, Herv\'e J\'egou, Julien Mairal,
  Piotr Bojanowski, and Armand Joulin.
\newblock Emerging properties in self-supervised vision transformers.
\newblock In \emph{Proceedings of the International Conference on Computer
  Vision (ICCV)}, 2021.

\bibitem[Chen et~al.(2020{\natexlab{a}})Chen, Radford, Child, Wu, Jun, Luan,
  and Sutskever]{chen2020generative}
Mark Chen, Alec Radford, Rewon Child, Jeffrey Wu, Heewoo Jun, David Luan, and
  Ilya Sutskever.
\newblock Generative pretraining from pixels.
\newblock In \emph{International conference on machine learning}, pages
  1691--1703. PMLR, 2020{\natexlab{a}}.

\bibitem[Chen et~al.(2020{\natexlab{b}})Chen, Kornblith, Norouzi, and
  Hinton]{chen2020simclr}
Ting Chen, Simon Kornblith, Mohammad Norouzi, and Geoffrey~E. Hinton.
\newblock A simple framework for contrastive learning of visual
  representations.
\newblock In \emph{International Conference on Machine Learning, {ICML}},
  2020{\natexlab{b}}.

\bibitem[Chen et~al.(2020{\natexlab{c}})Chen, Kornblith, Swersky, Norouzi, and
  Hinton]{chen2020big}
Ting Chen, Simon Kornblith, Kevin Swersky, Mohammad Norouzi, and Geoffrey~E
  Hinton.
\newblock Big self-supervised models are strong semi-supervised learners.
\newblock In \emph{Advances in Neural Information Processing Systems},
  2020{\natexlab{c}}.

\bibitem[Chen and He(2021)]{chen2021exploring}
Xinlei Chen and Kaiming He.
\newblock Exploring simple siamese representation learning.
\newblock In \emph{Proceedings of the IEEE/CVF Conference on Computer Vision
  and Pattern Recognition}, pages 15750--15758, 2021.

\bibitem[Chen et~al.(2020{\natexlab{d}})Chen, Fan, Girshick, and
  He]{chen2020improved}
Xinlei Chen, Haoqi Fan, Ross Girshick, and Kaiming He.
\newblock Improved baselines with momentum contrastive learning.
\newblock In \emph{arXiv preprint arXiv:2003.04297}, 2020{\natexlab{d}}.

\bibitem[Chen et~al.(2021)Chen, Xie, and He]{chen2021empirical}
Xinlei Chen, Saining Xie, and Kaiming He.
\newblock An empirical study of training self-supervised vision transformers.
\newblock In \emph{Proceedings of the IEEE/CVF international conference on
  computer vision}, pages 9640--9649, 2021.

\bibitem[Chen et~al.(2024)Chen, Ding, Wang, Xin, Mo, Wang, Han, Luo, Zeng, and
  Wang]{chen2024context}
Xiaokang Chen, Mingyu Ding, Xiaodi Wang, Ying Xin, Shentong Mo, Yunhao Wang,
  Shumin Han, Ping Luo, Gang Zeng, and Jingdong Wang.
\newblock Context autoencoder for self-supervised representation learning.
\newblock \emph{International Journal of Computer Vision}, 132\penalty0
  (1):\penalty0 208--223, 2024.

\bibitem[Christie et~al.(2018)Christie, Fendley, Wilson, and
  Mukherjee]{christie2018functional}
Gordon Christie, Neil Fendley, James Wilson, and Ryan Mukherjee.
\newblock Functional map of the world.
\newblock In \emph{Proceedings of the IEEE Conference on Computer Vision and
  Pattern Recognition}, pages 6172--6180, 2018.

\bibitem[Cimpoi et~al.(2014)Cimpoi, Maji, Kokkinos, Mohamed, and
  Vedaldi]{cimpoi2014describing}
Mircea Cimpoi, Subhransu Maji, Iasonas Kokkinos, Sammy Mohamed, and Andrea
  Vedaldi.
\newblock Describing textures in the wild.
\newblock In \emph{Proceedings of the IEEE conference on computer vision and
  pattern recognition}, pages 3606--3613, 2014.

\bibitem[Deng et~al.(2009)Deng, Dong, Socher, Li, Li, and
  Fei-Fei]{deng2009imagenet}
Jia Deng, Wei Dong, Richard Socher, Li-Jia Li, Kai Li, and Li Fei-Fei.
\newblock Imagenet: A large-scale hierarchical image database.
\newblock In \emph{2009 IEEE conference on computer vision and pattern
  recognition}, pages 248--255. Ieee, 2009.

\bibitem[Dosovitskiy et~al.(2014)Dosovitskiy, Springenberg, Riedmiller, and
  Brox]{dosovitskiy2014discriminative}
Alexey Dosovitskiy, Jost~Tobias Springenberg, Martin Riedmiller, and Thomas
  Brox.
\newblock Discriminative unsupervised feature learning with convolutional
  neural networks.
\newblock In \emph{Advances in neural information processing systems}, pages
  766--774, 2014.

\bibitem[Dosovitskiy et~al.(2021)Dosovitskiy, Beyer, Kolesnikov, Weissenborn,
  Zhai, Unterthiner, Dehghani, Minderer, Heigold, Gelly, Uszkoreit, and
  Houlsby]{dosovitskiy2021an}
Alexey Dosovitskiy, Lucas Beyer, Alexander Kolesnikov, Dirk Weissenborn,
  Xiaohua Zhai, Thomas Unterthiner, Mostafa Dehghani, Matthias Minderer, Georg
  Heigold, Sylvain Gelly, Jakob Uszkoreit, and Neil Houlsby.
\newblock An image is worth 16x16 words: Transformers for image recognition at
  scale.
\newblock In \emph{International Conference on Learning Representations}, 2021.

\bibitem[Dubey et~al.(2024)Dubey, Jauhri, Pandey, Kadian, Al-Dahle, Letman,
  Mathur, Schelten, Yang, Fan, et~al.]{dubey2024llama3}
Abhimanyu Dubey, Abhinav Jauhri, Abhinav Pandey, Abhishek Kadian, Ahmad
  Al-Dahle, Aiesha Letman, Akhil Mathur, Alan Schelten, Amy Yang, Angela Fan,
  et~al.
\newblock The llama 3 herd of models.
\newblock \emph{arXiv preprint arXiv:2407.21783}, 2024.

\bibitem[El-Nouby et~al.(2024)El-Nouby, Klein, Zhai, Bautista, Toshev, Shankar,
  Susskind, and Joulin]{el2024scalable}
Alaaeldin El-Nouby, Michal Klein, Shuangfei Zhai, Miguel~Angel Bautista,
  Alexander Toshev, Vaishaal Shankar, Joshua~M Susskind, and Armand Joulin.
\newblock Scalable pre-training of large autoregressive image models.
\newblock In \emph{International conference on machine learning}. PMLR, 2024.

\bibitem[Grill et~al.(2020)Grill, Strub, Altch\'{e}, Tallec, Richemond,
  Buchatskaya, Doersch, Avila~Pires, Guo, Gheshlaghi~Azar, Piot, kavukcuoglu,
  Munos, and Valko]{grill2020bootstrap}
Jean-Bastien Grill, Florian Strub, Florent Altch\'{e}, Corentin Tallec, Pierre
  Richemond, Elena Buchatskaya, Carl Doersch, Bernardo Avila~Pires, Zhaohan
  Guo, Mohammad Gheshlaghi~Azar, Bilal Piot, koray kavukcuoglu, Remi Munos, and
  Michal Valko.
\newblock Bootstrap your own latent - a new approach to self-supervised
  learning.
\newblock In \emph{Advances in Neural Information Processing Systems}, 2020.

\bibitem[He et~al.(2020)He, Fan, Wu, Xie, and Girshick]{he2020momentum}
Kaiming He, Haoqi Fan, Yuxin Wu, Saining Xie, and Ross Girshick.
\newblock Momentum contrast for unsupervised visual representation learning.
\newblock In \emph{Proceedings of the IEEE/CVF Conference on Computer Vision
  and Pattern Recognition}, pages 9729--9738, 2020.

\bibitem[He et~al.(2022)He, Chen, Xie, Li, Doll{\'a}r, and
  Girshick]{he2022masked}
Kaiming He, Xinlei Chen, Saining Xie, Yanghao Li, Piotr Doll{\'a}r, and Ross
  Girshick.
\newblock Masked autoencoders are scalable vision learners.
\newblock In \emph{Proceedings of the IEEE/CVF conference on computer vision
  and pattern recognition}, pages 16000--16009, 2022.

\bibitem[Helber et~al.(2019)Helber, Bischke, Dengel, and
  Borth]{helber2019eurosat}
Patrick Helber, Benjamin Bischke, Andreas Dengel, and Damian Borth.
\newblock Eurosat: A novel dataset and deep learning benchmark for land use and
  land cover classification.
\newblock \emph{IEEE Journal of Selected Topics in Applied Earth Observations
  and Remote Sensing}, 12\penalty0 (7):\penalty0 2217--2226, 2019.

\bibitem[Kaplan et~al.(2020)Kaplan, McCandlish, Henighan, Brown, Chess, Child,
  Gray, Radford, Wu, and Amodei]{kaplan2020scaling}
Jared Kaplan, Sam McCandlish, Tom Henighan, Tom~B Brown, Benjamin Chess, Rewon
  Child, Scott Gray, Alec Radford, Jeffrey Wu, and Dario Amodei.
\newblock Scaling laws for neural language models.
\newblock \emph{arXiv preprint arXiv:2001.08361}, 2020.

\bibitem[Kenton and Toutanova(2019)]{kenton2019bert}
Jacob Devlin Ming-Wei~Chang Kenton and Lee~Kristina Toutanova.
\newblock Bert: Pre-training of deep bidirectional transformers for language
  understanding.
\newblock In \emph{Proceedings of naacL-HLT}, page~2. Minneapolis, Minnesota,
  2019.

\bibitem[Krause et~al.(2013)Krause, Stark, Deng, and Fei-Fei]{krause20133d}
Jonathan Krause, Michael Stark, Jia Deng, and Li Fei-Fei.
\newblock 3d object representations for fine-grained categorization.
\newblock In \emph{Proceedings of the IEEE international conference on computer
  vision workshops}, pages 554--561, 2013.

\bibitem[Krizhevsky et~al.(2009)Krizhevsky, Hinton,
  et~al.]{krizhevsky2009learning}
Alex Krizhevsky, Geoffrey Hinton, et~al.
\newblock Learning multiple layers of features from tiny images.
\newblock Technical report, University of Toronto, 2009.

\bibitem[Oquab et~al.(2023)Oquab, Darcet, Moutakanni, Vo, Szafraniec, Khalidov,
  Fernandez, Haziza, Massa, El-Nouby, et~al.]{oquab2023dinov2}
Maxime Oquab, Timoth{\'e}e Darcet, Th{\'e}o Moutakanni, Huy Vo, Marc
  Szafraniec, Vasil Khalidov, Pierre Fernandez, Daniel Haziza, Francisco Massa,
  Alaaeldin El-Nouby, et~al.
\newblock Dinov2: Learning robust visual features without supervision.
\newblock \emph{arXiv preprint arXiv:2304.07193}, 2023.

\bibitem[Parkhi et~al.(2012)Parkhi, Vedaldi, Zisserman, and
  Jawahar]{parkhi2012cats}
Omkar~M Parkhi, Andrea Vedaldi, Andrew Zisserman, and CV Jawahar.
\newblock Cats and dogs.
\newblock In \emph{2012 IEEE conference on computer vision and pattern
  recognition}, pages 3498--3505. IEEE, 2012.

\bibitem[Peng et~al.(2019)Peng, Bai, Xia, Huang, Saenko, and
  Wang]{peng2019moment}
Xingchao Peng, Qinxun Bai, Xide Xia, Zijun Huang, Kate Saenko, and Bo Wang.
\newblock Moment matching for multi-source domain adaptation.
\newblock In \emph{Proceedings of the IEEE/CVF international conference on
  computer vision}, pages 1406--1415, 2019.

\bibitem[Radford et~al.(2019)Radford, Wu, Child, Luan, Amodei, Sutskever,
  et~al.]{radford2019language}
Alec Radford, Jeffrey Wu, Rewon Child, David Luan, Dario Amodei, Ilya
  Sutskever, et~al.
\newblock Language models are unsupervised multitask learners.
\newblock \emph{OpenAI blog}, 1\penalty0 (8):\penalty0 9, 2019.

\bibitem[Rolfe(2016)]{rolfe2016discrete}
Jason~Tyler Rolfe.
\newblock Discrete variational autoencoders.
\newblock \emph{arXiv preprint arXiv:1609.02200}, 2016.

\bibitem[Sablayrolles et~al.(2018)Sablayrolles, Douze, Schmid, and
  J{\'e}gou]{sablayrolles2018spreading}
Alexandre Sablayrolles, Matthijs Douze, Cordelia Schmid, and Herv{\'e}
  J{\'e}gou.
\newblock Spreading vectors for similarity search.
\newblock \emph{arXiv preprint arXiv:1806.03198}, 2018.

\bibitem[Singh et~al.(2023)Singh, Duval, Alwala, Fan, Aggarwal, Adcock, Joulin,
  Doll{\'a}r, Feichtenhofer, Girshick, et~al.]{singh2023effectiveness}
Mannat Singh, Quentin Duval, Kalyan~Vasudev Alwala, Haoqi Fan, Vaibhav
  Aggarwal, Aaron Adcock, Armand Joulin, Piotr Doll{\'a}r, Christoph
  Feichtenhofer, Ross Girshick, et~al.
\newblock The effectiveness of mae pre-pretraining for billion-scale
  pretraining.
\newblock In \emph{Proceedings of the IEEE/CVF International Conference on
  Computer Vision}, pages 5484--5494, 2023.

\bibitem[Taylor et~al.(2019)Taylor, Earnshaw, Mabey, Victors, and
  Yosinski]{taylor2019rxrx1}
James Taylor, Berton Earnshaw, Ben Mabey, Mason Victors, and Jason Yosinski.
\newblock Rxrx1: An image set for cellular morphological variation across many
  experimental batches.
\newblock In \emph{International Conference on Learning Representations
  (ICLR)}, page~23, 2019.

\bibitem[Touvron et~al.(2021)Touvron, Cord, Sablayrolles, Synnaeve, and
  J{\'e}gou]{touvron2021going}
Hugo Touvron, Matthieu Cord, Alexandre Sablayrolles, Gabriel Synnaeve, and
  Herv{\'e} J{\'e}gou.
\newblock Going deeper with image transformers.
\newblock In \emph{Proceedings of the IEEE/CVF international conference on
  computer vision}, pages 32--42, 2021.

\bibitem[Touvron et~al.(2023{\natexlab{a}})Touvron, Lavril, Izacard, Martinet,
  Lachaux, Lacroix, Rozi{\`e}re, Goyal, Hambro, Azhar,
  et~al.]{touvron2023llama}
Hugo Touvron, Thibaut Lavril, Gautier Izacard, Xavier Martinet, Marie-Anne
  Lachaux, Timoth{\'e}e Lacroix, Baptiste Rozi{\`e}re, Naman Goyal, Eric
  Hambro, Faisal Azhar, et~al.
\newblock Llama: Open and efficient foundation language models.
\newblock \emph{arXiv preprint arXiv:2302.13971}, 2023{\natexlab{a}}.

\bibitem[Touvron et~al.(2023{\natexlab{b}})Touvron, Martin, Stone, Albert,
  Almahairi, Babaei, Bashlykov, Batra, Bhargava, Bhosale,
  et~al.]{touvron2023llama2}
Hugo Touvron, Louis Martin, Kevin Stone, Peter Albert, Amjad Almahairi, Yasmine
  Babaei, Nikolay Bashlykov, Soumya Batra, Prajjwal Bhargava, Shruti Bhosale,
  et~al.
\newblock Llama 2: Open foundation and fine-tuned chat models.
\newblock \emph{arXiv preprint arXiv:2307.09288}, 2023{\natexlab{b}}.

\bibitem[Van~Horn et~al.(2018)Van~Horn, Mac~Aodha, Cui, Song, Shepard, Adam,
  Perona, and Belongie]{inaturalist18}
Grant Van~Horn, Oisin Mac~Aodha, Yin Cui, Yang Song, Alex Shepard, Hartwig
  Adam, Pietro Perona, and Serge Belongie.
\newblock {iNaturalist} 2018 competition dataset.
\newblock \url{https://github.com/visipedia/inat_comp/tree/master/2018}, 2018.

\bibitem[Veeling et~al.(2018)Veeling, Linmans, Winkens, Cohen, and
  Welling]{veeling2018rotation}
Bastiaan~S Veeling, Jasper Linmans, Jim Winkens, Taco Cohen, and Max Welling.
\newblock Rotation equivariant cnns for digital pathology.
\newblock In \emph{Medical Image Computing and Computer Assisted
  Intervention--MICCAI 2018: 21st International Conference, Granada, Spain,
  September 16-20, 2018, Proceedings, Part II 11}, pages 210--218. Springer,
  2018.

\bibitem[Wu et~al.(2018)Wu, Xiong, Yu, and Lin]{wu2018unsupervised}
Zhirong Wu, Yuanjun Xiong, Stella~X Yu, and Dahua Lin.
\newblock Unsupervised feature learning via non-parametric instance
  discrimination.
\newblock In \emph{Proceedings of the IEEE Conference on Computer Vision and
  Pattern Recognition}, pages 3733--3742, 2018.

\bibitem[Xie et~al.(2022)Xie, Zhang, Cao, Lin, Bao, Yao, Dai, and
  Hu]{xie2022simmim}
Zhenda Xie, Zheng Zhang, Yue Cao, Yutong Lin, Jianmin Bao, Zhuliang Yao, Qi
  Dai, and Han Hu.
\newblock Simmim: A simple framework for masked image modeling.
\newblock In \emph{Proceedings of the IEEE/CVF conference on computer vision
  and pattern recognition}, pages 9653--9663, 2022.

\bibitem[Zhou et~al.(2021)Zhou, Wei, Wang, Shen, Xie, Yuille, and
  Kong]{zhou2021ibot}
Jinghao Zhou, Chen Wei, Huiyu Wang, Wei Shen, Cihang Xie, Alan Yuille, and Tao
  Kong.
\newblock ibot: Image bert pre-training with online tokenizer.
\newblock \emph{arXiv preprint arXiv:2111.07832}, 2021.

\end{thebibliography}
